\documentclass[runningheads]{llncs}

\usepackage{eccv}

\usepackage{eccvabbrv}

\usepackage{graphicx}
\usepackage{booktabs}

\usepackage[accsupp]{axessibility}  

\usepackage{hyperref}

\usepackage{orcidlink}

\usepackage{multirow}
\usepackage{array}

\begin{document}

\title{
Explanation for Trajectory Planning using 
Multi-modal Large Language Model 
for Autonomous Driving
}

\titlerunning{Explanation for Trajectory Planning using MLLM for AD}

\author{Shota Yamazaki\inst{1}\orcidlink{0009-0004-9226-099X} \and
Chenyu Zhang\inst{1} \and
Takuya Nanri\inst{1} \and
Akio Shigekane\inst{1} \and
\\
Siyuan Wang\inst{1} \and
Jo Nishiyama\inst{1} \and
Tao Chu\inst{1} \and
Kohei Yokosawa\inst{2}
}

\authorrunning{S.~Yamazaki et al.}

\institute{
Mobility and AI Laboratory, Nissan Motor Co., Ltd.
\and
Prototype and Test Department, Nissan Motor Co., Ltd.\\1-1-1, Takashima, Nishi-ku, Yokohama-shi, Kanagawa, 220-8686, Japan
}

\maketitle

\begin{abstract}

End-to-end style autonomous driving models have been developed recently. 
These models lack interpretability of decision-making process from perception to control of the ego vehicle, resulting in anxiety for passengers.
To alleviate it, it is effective to build a model which outputs captions describing future behaviors of the ego vehicle and their reason. 
However, the existing approaches generate reasoning text that inadequately reflects the future plans of the ego vehicle, because they train models to output captions using momentary control signals as inputs.
In this study, we propose a reasoning model that takes future planning trajectories of the ego vehicle as inputs to solve this limitation with the dataset newly collected.

\end{abstract}

\section{Introduction}
\label{sec:intro}

\begin{figure}[tb]
    \centering
    \includegraphics[width=\linewidth]{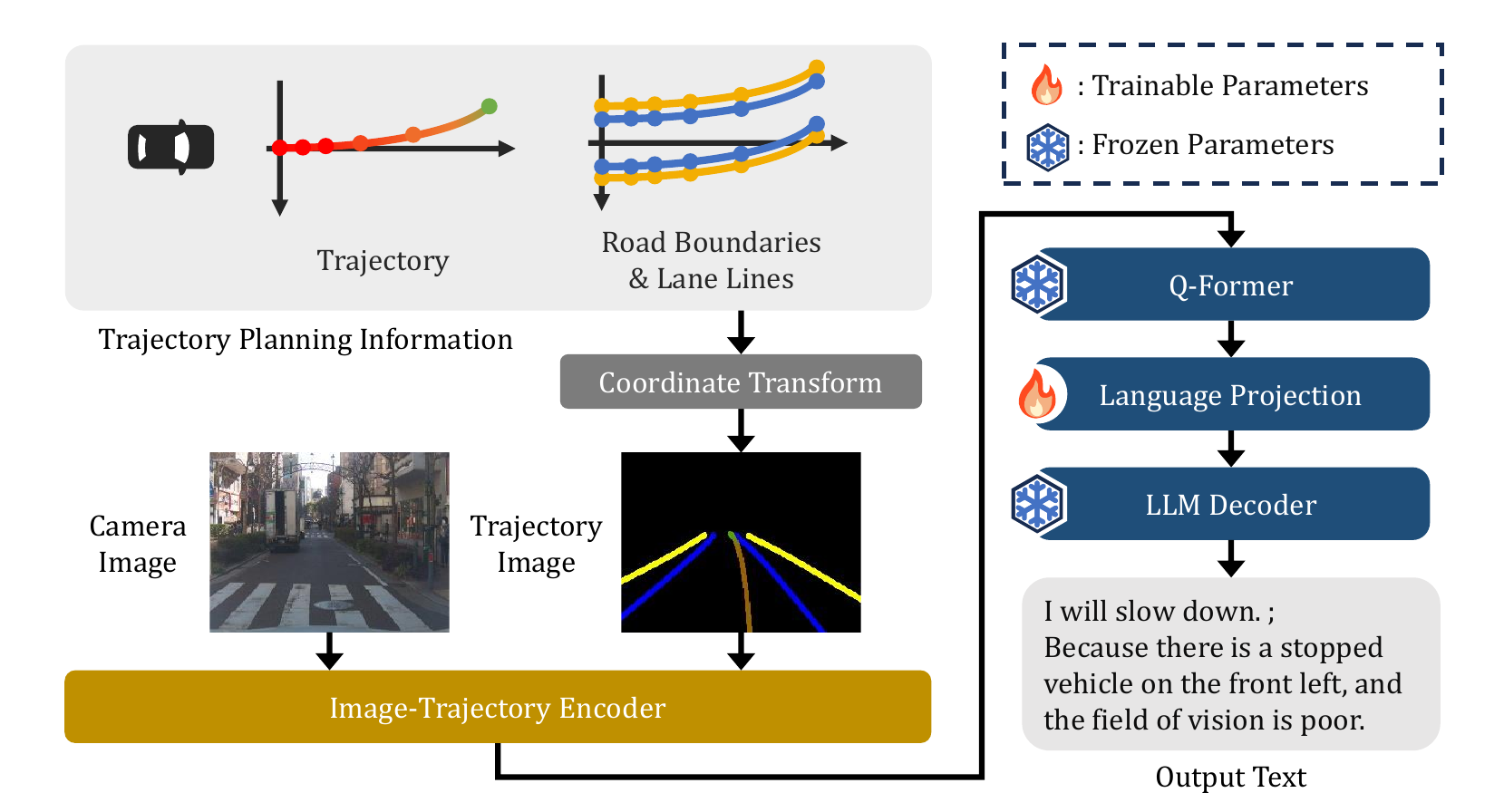}
    \caption{Pipeline of our proposed method. In order to improve accountability of an ego-vehicle action, trajectory planning information is embedded as a trajectory image and combined to a camera image in a Image-Trajectory Encoder.}
    \label{fig:pipeline}
\end{figure}

Research on autonomous driving has been actively conducted since 2010s after the Darpa Grand Challenge \cite{10.5555/1554715}. The advancement of deep learning has significantly improved performance of machine learning tasks such as object detection and semantic segmentation, and has contributed to autonomous driving research.
A typical architecture of a autonomous driving system is a modular system, which consists of separate components such as localization, perception, prediction, planning, and control \cite{9046805}. 
On the other hand, an end-to-end approach\cite{e2esurvey}, which integrates all components and performs from inputs of sensor data to outputs of control signals consistently, has been emerged recently\cite{drivegpt4}.
It has expected to avoid cumulative errors occurred in each component of the modular system, but it is difficult to know the process of decision making in the approach. Passengers feel anxiety when they cannot understand what the ego vehicle recognizes and why it takes the action.
Recent researches have tried to add a large language model (LLM) such as GPT-4\cite{openai2024gpt4technicalreport} and generate captions describing behaviors of the ego vehicle and reasons for them in order to solve the above problem\cite{adapt, drivegpt4, bddx, drivelm, nuscenesqa, 10030940, 10483594}.
However, the existing methods have a limitation that they can describe only current or past actions. 

In this paper, we propose a method, in which a visual image and future driving plans (trajectory planning information) are combined and their fused features are used to generate accurate captions of actions which the ego vehicle will take and reasons for them.
To this end, we collect and build a new dataset which includes both trajectory planning information and its captions which the existing datasets do not have, because where the ego vehicle should pay attention depends on the trajectory planning information.
Let us consider the scenario in which a vehicle ahead of the ego vehicle is stopping.
If the ego vehicle wants to stop in front of the vehicle ahead, it will confirm the space between them. Conversely, if the intention of the ego vehicle is to proceed by avoiding the front vehicle to the left and right sides of the road, it should pay attention to the surrounding situations, including the presence of pedestrians and vehicles in adjacent lanes.

The main contributions of this paper are summarized as follows:
\begin {itemize}
    \item We present a new approach to the spatial fusion of visual information and trajectory planning information using cross attention.
    \item We demonstrate improvement of generated captions that describe and justify future behaviors of the ego vehicle, 
    using the fused features and a BLIP-2\cite{blip2} based vision-language model.
    \item We compile and annotate a new dataset consisting of videos, trajectory planning information, and captions for them.
 \end {itemize}

\section{Related Works}
\label{sec:related}

This section describes related studies of caption generation as well as studies on ego motion caption datasets.

\subsection{Caption Generation}
Caption generation is a task to generate text describing an input image. With the development of deep learning, some methods using LSTM \cite {lstm} such as Show and Tell \cite{showandtell} and Neural Baby Talk\cite{nbt} have been proposed. In recent years, Transformer\cite{NIPS2017_3f5ee243} and the growing size of datasets have dramatically improved performance of various tasks.
BLIP-2 \cite{blip2} is a pioneering vision-language model for caption generation that is the basis of several applied studies. BLIP-2 leverages and combines the existing vision encoders and LLM models via a transformer-based network called Q-Former to achieve image caption generation and visual question answering. BLIP-2 has succeeded in reduction of the cost of model training and maintenance of the performance whereas only the Q-Former is trained.

Some studies have applied vision-language models for generation of captions about driving behaviors of the ego vehicle. ADAPT \cite{adapt} simultaneously optimized a module that predicts control signals and one that generates descriptions and justification of an action of the ego vehicle from video tokens by multi-task learning.
Other studies have used not only visual information but also control information as inputs to generate descriptions and justification of ego motions.
For example, DriveGPT4 \cite{drivegpt4} used a multi-modal LLM, which inputs video flames, text of a question, and past control variables, and outputs text of an answer and next control variables.
However, these studies have not considered planning information such as future trajectory path of the ego vehicle.

\subsection{Driving Action Caption Dataset}
BDD-X \cite{bddx} has been a pioneering study that describes and justifies driving behaviors of the ego vehicle. BDD-X built a dataset consisting of approximately 40 seconds of videos and driving activities, and 7,000 textual descriptions of ego-vehicle actions and their justifications. For example, a description of an ego motion can be "The car moves back into the left lane", and its justification can be "because the school bus in front of it is stopping". ADAPT and DriveGPT4 are examples which used BDD-X.

Recent researches have described driving scenes more comprehensively. 
For example, DriveLM \cite{drivelm} described questions and answers (QAs) for objects, situations, and control instructions for the ego vehicle in text. 
DriveLM added text with three-dimensional annotations to a dataset nuScenes \cite{nuscenes}. 
NuScenes-QA \cite {nuscenesqa} described QAs for objects.
DRAMA \cite{10030940} also annotated objects and provides captions of surrounding circumstances and suggested actions. Moreover, Rank2Tell \cite{10483594} added risk level of each object to identify important objects which affect ego motions.
Many studies have recently been conducted to generate captions in nuScenes such as adding text to it. However, these studies have been concerned only with relations between objects and situations in a scene, and between objects and current control instructions.
In contrast to the existing datasets above, our new dataset focuses on situations and future plans of ego-vehicle behaviors.

\section{Method}
\subsection{Overview}

\cref{fig:pipeline} shows the pipeline of the proposed method, which is based on BLIP-2 \cite{blip2}.
It consists of a coordinate transformation module of trajectory planning information, an Image-Trajectory Encoder, a Q-Former, a language projection module, and a LLM encoder.
Trajectory planning information includes a future trajectory of the ego vehicle, a pair of road boundaries, and a pair of lane lines.
It is converted into a trajectory image in a coordinate transformation module.
In the Image-Trajectory Encoder, an front camera image and the trajectory image of the ego vehicle are input, and their fused features are extracted.
The extracted features are passed to the Q-Former and the language projection module of BLIP-2, and finally, the LLM decoder outputs explanations and justifications of the action of the ego vehicle.

\subsection{Trajectory Planning Information}
\label{subsec:traj}
A planned future trajectory of the ego vehicle \(P_i\), a road boundary \(B_i\), and a lane line \(L_i\) on the \(i\)-th time frame are generated from the autonomous driving system on the ego vehicle.
They consist of a group of spatial coordinates on the Cartesian coordinate system such as \(P_i = ((x_1, y_1, z_1),..., (x_N, y_N, z_N)); i = (1,..., T)\) where \(T\) represents the total frame number and \(N\) represents the number of coordinates.
In DriveGPT4 \cite{drivegpt4}, trajectory planning information is treated as text information, which is input into a multi-modal LLM.
However, in our proposed pipeline, it is transformed from Cartesian coordinates to image space coordinates by using perspective projection transformation after translation and rotation operations.
In addition, the points on the image space coordinate system are connected and lines are drawn as a trajectory image, which has a same angle of view as a front camera image.

Herein, each line is drawn in a different color. For examples, a pair of road boundaries is drawn in yellow, and a pair of lane lines is in blue. Furthermore, in order to keep the velocity information of the trajectory, the color of the trajectory line changes with the velocity of the ego vehicle; fast in red and slow in green.

\subsection{Image-Trajectory Encoder}
\label{subsec:image-trajectory_encoder}

\begin{figure}[tb]
  \centering

  \begin{subfigure}{0.32\linewidth}
    \centering
    \includegraphics[height=6.1cm]{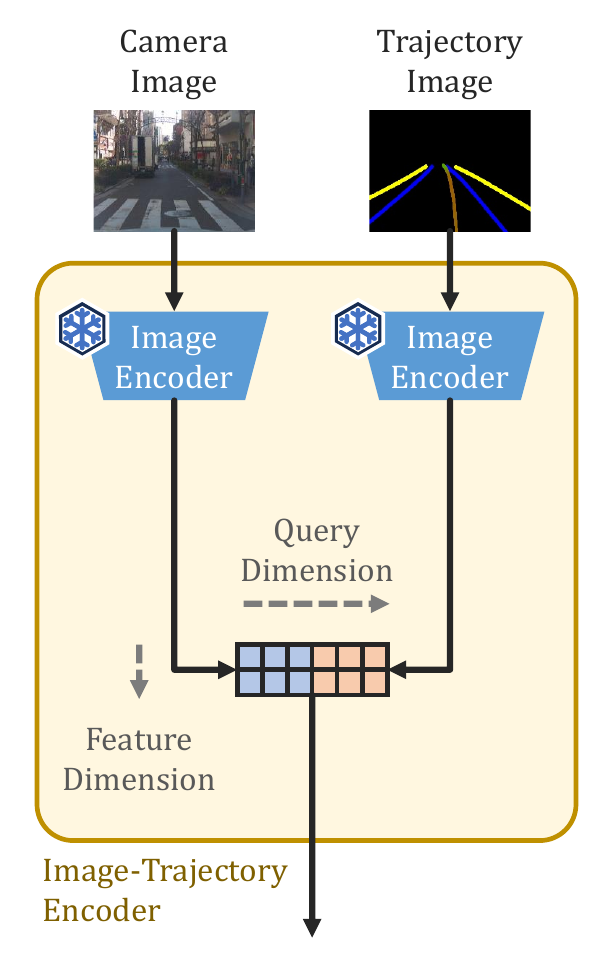}
    \caption{Concatenated}
    \label{fig:concat}
  \end{subfigure}
  \begin{subfigure}{0.26\linewidth}
    \centering
    \includegraphics[height=6.1cm]{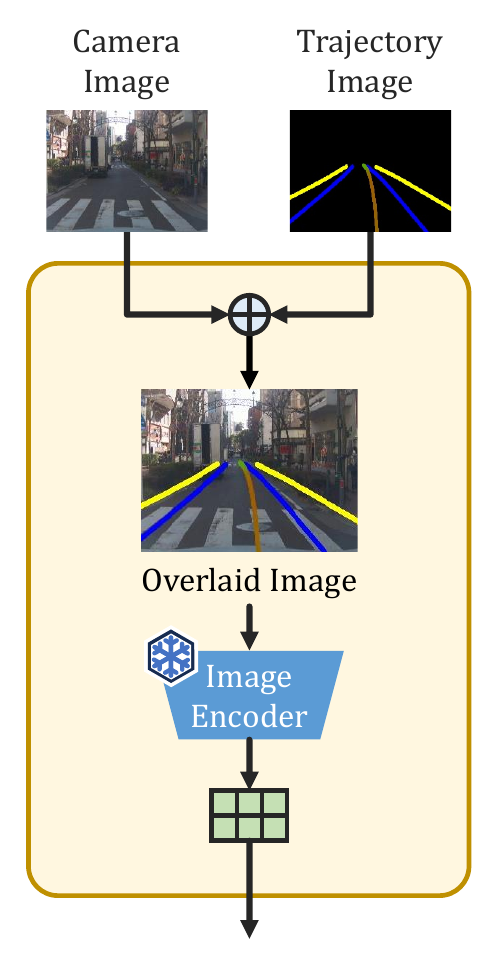}
    \caption{Overlaid}
    \label{fig:overlaid}
  \end{subfigure}
  \begin{subfigure}{0.37\linewidth}
    \centering
    \includegraphics[height=6.1cm]{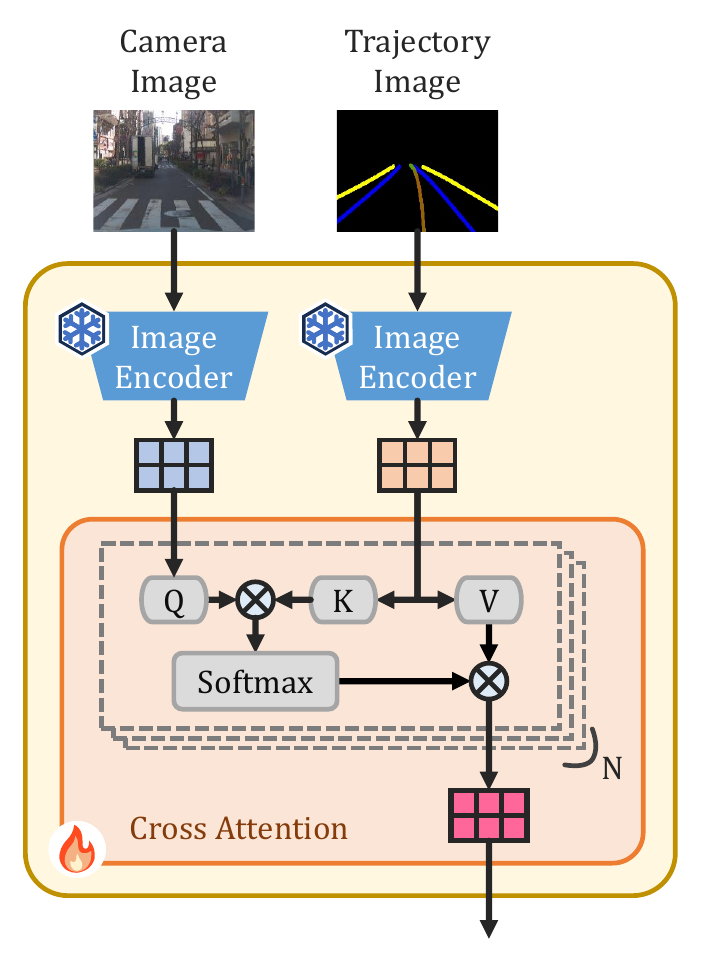}
    \caption{Cross-attention}
    \label{fig:cross-attention}
  \end{subfigure}
  \caption{Architectures of Image-Trajectory Encoders. (a) Concatenated. Both front camera image features and trajectory features are simply concatenated. (b) Overlaid. The trajectory image is overlaid on the camera image. (c) Cross-attention. With the camera image features as queries \(Q\) and the trajectory features as keys \(K\) and values \(V\), cross attention layers extract fused features.}
  \label{fig:image-trajectory_encoder}

\end{figure}

It is important how to connect trajectory planning information with visual information of a front camera image.
Here we consider three types of architecture for Image-Trajectory Encoders as shown in \cref{fig:image-trajectory_encoder}.

\subsubsection{Concatenated.}
This architecture has two image encoders, which are based on BLIP-2.
One extracts features from an front camera image, and the other does from an trajectory image.
The two pairs of the features are concatenated and output.
In our experiments, each image encoder outputs 257 queries of features where each query has a dimension of 1408. After the concatenation, the size of the final features is \(514 \times 1408\).
This method can simply pass trajectory planning information to subsequent modules.

\subsubsection{Overlaid.}
In the overlaid method, a trajectory image is overlaid on a front camera image. The overlaid image is input to a image encoder and combined features of the visual information and the trajectory planning information are extracted.
This method intends to clarify the spatial relations between the camera image and the trajectory.
As the trajectory planning information is directly embedded to the image, we do not have to modify the architecture of the original BLIP-2 model.

\subsubsection{Cross-attention.}
This method extracts features from two image encoders as well as the concatenated, but the way to fuse features of a front camera image and a trajectory image is different. There are cross-attention layers in this architecture. The features of the visual information are input to the layers as queries, whereas the features of the trajectory planning information are as keys and values. As a result, the both features interact with one another and the fused features are acquired.

\section{Experiment}

\subsection{Model Implementation}
We compare the performance of the proposed models described in \cref{subsec:image-trajectory_encoder} with that of a baseline model, which has the same architecture as the original BLIP-2 model and generates captions from only a front camera image.
All models are implemented in Transformers \cite{wolf-etal-2020-transformers} and utilize the pretrained model Salesforce/blip2-opt-2.7b from Hugging Face \cite{salesforce-blip2-pretrained}.

The number of training epochs is approximately 10 for each method. As fine-tuning of a CLIP-based visual encoder needs cautious alignment, parameters of image encoders, Q-Formers, and LLM decoders are frozen during a training stage.

\subsection{Dataset}
\begin{figure}[tb]
  \centering

  \begin{tabular}{cccc}
    \includegraphics[width=0.23\linewidth, height=0.1725\linewidth]{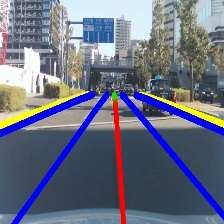} &
    \includegraphics[width=0.23\linewidth, height=0.1725\linewidth]{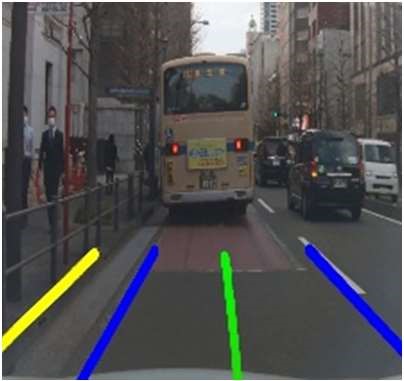} &
    \includegraphics[width=0.23\linewidth, height=0.1725\linewidth]{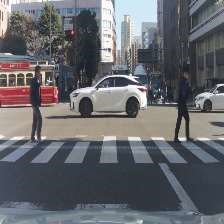} &
    \includegraphics[width=0.23\linewidth, height=0.1725\linewidth]{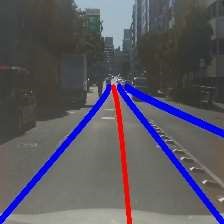} \\
    (a) & (b) & (c) & (d) \\
  \end{tabular}

  \begin{tabular}{cp{0.94\linewidth}}
    &\\
    (a) & I will drive at a steady speed. ; Because there is a safe distance from the front vehicle.\\
    (b) & I will slow down. ; Because the front vehicle is stopped.\\
    (c) & I will maintain the parked state. ; Because the traffic light is red.\\
    (d) & I will drive at a steady speed. ; Because there is a safe distance from the front vehicle.\\
  \end{tabular}

  \caption{Examples of our dedicated dataset}
  \label{fig:dataset}
\end{figure}

Future planning trajectories and captions on the ego motion are required for the dataset in our experiment. To make the dataset, we collected driving data, including future trajectories, using an autonomous driving system. Furthermore, we annotated the driving data with the captions.

We adopted comma 3X developed by Comma.ai for the hardware implementation to generate the future trajectories, as it is a commercial product and makes it easier to replicate our experiments.
The comma 3X was installed at the front of the ego vehicle, and front camera images and future trajectories were simultaneously acquired during driving.
The recording was done in the urban district of Japan and approximately 120 min of recorded driving data was used for the dataset. It consists of 69 min of the data on main roads where vehicles mainly drive, and 51 min on narrow streets where pedestrians may cross.
Moreover, explanation and justification of the action were manually annotated in English for each frame. \cref{fig:dataset} shows examples of our dedicated dataset.

\subsection{Quantitative Evaluation}

\begin{table}[tb]
    \caption{Quantitative evaluation results of four methods on two metrics. B4: BLEU-4, RL: ROUGE-L.}
    \label{tab:evaluation_results}
    \centering
    \begin{tabular}{lcccccc}
        \toprule
        \multicolumn{1}{c}{\multirow{2}{*}{Model}} & \multicolumn{2}{c}{Whole} & \multicolumn{2}{c}{Action} & \multicolumn{2}{c}{Justification} \\
        \cmidrule(lr){2-3}
        \cmidrule(lr){4-5}
        \cmidrule(lr){6-7}
            & B4\(\uparrow\) & RL\(\uparrow\) & B4\(\uparrow\) & RL\(\uparrow\) & B4\(\uparrow\) & RL\(\uparrow\) \\
        \midrule
        Baseline & 0.337 & 0.524 & 0.361 & 0.547 & 0.235 & 0.510 \\ \midrule 
        Concatenated& 0.378& 0.548& 0.417& 0.590& 0.290& 0.526\\
        Overlaid & 0.379 & 0.547 & 0.478 & 0.626 & 0.277 & 0.502 \\
        \textbf{Cross-attention}& \textbf{0.416}& \textbf{0.566}& \textbf{0.483}& \textbf{0.627}& \textbf{0.340}& \textbf{0.535}\\
        \bottomrule
        \end{tabular}
\end{table}

We train and test the performance of the proposed models on our dataset. The captioning performance is evaluated using BLEU-4 \cite{10.3115/1073083.1073135} and ROUGE-L \cite{lin-2004-rouge}, which are general metrics of caption generation tasks. Each metric aims to measure a distinct aspect of the generated sentences: BLEU-4 assesses their precision, whereas ROUGE-L evaluates their recall.
The evaluations are not only for the whole sentences, but also for the sentences divided into actions and justifications.

\cref{tab:evaluation_results} shows the quantitative results of the baseline and proposed models.
Both BLEU-4 and ROUGE-L are improved in all of the proposed models compared to the baseline, that implies a positive effect of adding trajectory planning information to the model inputs.
The overlaid model performs better in the action explanation, while the justification performance becomes worse than the concatenated model.
The reason for this can be explained as follows. The overlaid model can explicitly combine a camera image and trajectory planning information considering the spatial relations between them, that contributes better action explanations. On the other hand, it loses some information such as a front vehicle and a pedestrian on the camera image because of overlaid lines, which is needed for justification explanations.
The cross-attention model solves the above problem and achieves the best scores in both action and justification captions.

\subsection{Qualitative Evaluation}

\begin{figure}[tb]
  \centering
  \begin{tabular}{cc}
    \includegraphics[width=0.3\linewidth, height=0.225\linewidth]{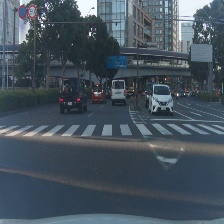} &
    \includegraphics[width=0.3\linewidth, height=0.225\linewidth]{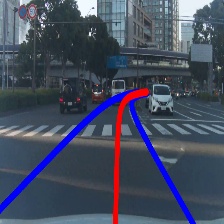} \\
    Original image & Overlaid image \\
  \end{tabular}

  \begin{tabular}{c|l}
    \toprule
    Ground Truth & I will drive at a steady speed. \\
        & Because there is a safe distance from the front vehicle. \\
    \hline
    Baseline & I will maintain the parked state. \\
        & Because the front vehicle is stopped. \\
    \hline
    Concatenated & I will accelerate gradually. \\
        & Because the front vehicle is accelerating gradually. \\
    \hline
    Overlaid & I will drive at a steady speed. \\
        & Because the front vehicle is stopped. \\
    \hline
    Cross-attention & I will drive at a steady speed. \\
        & Because there is a safe distance from the front vehicle. \\
    \bottomrule
  \end{tabular}
  
  \caption{Example of generated results}
  \label{fig:output_sample01}
\end{figure}

\begin{figure}[tb]
  \centering
  \begin{tabular}{cc}
    \includegraphics[width=0.3\linewidth, height=0.225\linewidth]{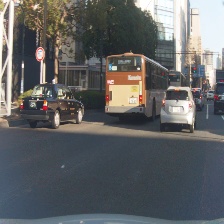} &
    \includegraphics[width=0.3\linewidth, height=0.225\linewidth]{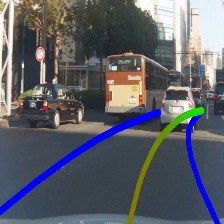} \\
    Original image & Overlaid image \\
  \end{tabular}

  \begin{tabular}{c|l}
    \toprule
    Ground Truth & I will slow down. \\
        & Because the front vehicle slowed down. \\
    \hline
    Baseline & I will maintain the parked state. \\
        & Because the front vehicle is stopped. \\
    \hline
    Concatenated & I will accelerate gradually. \\
        & Because the front vehicle is accelerating gradually. \\
    \hline
    Overlaid & I will accelerate gradually. \\
        & Because the front vehicle is accelerating gradually. \\
    \hline
    Cross-attention & I will accelerate gradually. \\
        & Because the front vehicle is accelerating gradually. \\
    \bottomrule
  \end{tabular}
  
  \caption{Limitation example of generated results}
  \label{fig:output_sample02}
\end{figure}

\cref{fig:output_sample01} shows the difference between the results of the baseline model and that of the proposed models.
In the scene shown in \cref{fig:output_sample01}, there is enough space in front of the ego vehicle and it is driving straight forward at a steady speed.
The baseline model cannot understand the current status of the ego vehicle, and the wrong explanation of the action is described.
Both the concatenated model and the overlaid model understand that the ego vehicle is going ahead, but cannot fully capture the relation between its behavior and surrounding situations.
Only the cross-attention model succeeds in explaining the action and justification of the ego vehicle precisely.

While \cref{fig:output_sample01} shows effectiveness of our proposed methods, there are some limitations.
In the scene shown in \cref{fig:output_sample02}, the ego vehicle is moving straight ahead diagonally to the right, and is decelerating due to the congestion ahead.
However, none of the baseline model or our proposed models cannot correctly capture the behavior of the front vehicle and incorrectly explains that it is accelerating or stopping.
The reason is that it is difficult for the models to estimate a spatio-temporal behavior of an object from a single image without temporal information.

\subsection{Additional Ablation Study on Cross-attention Layers}
Our proposed cross-attention model uses
a front camera image as queries and trajectory planning information as keys and values as described in \cref{subsec:image-trajectory_encoder}, but there are some discussions on which features should be queries, keys, or values.

BLOS-BEV \cite{blos} fused a SD map and visual BEV features, and achieved great results by utilizing the SD map features as queries and the visual BEV features as keys and values for cross attention calculation. In this paper, the cross attention structure with the similar concept is used. Thus we would like to explore how different selections of queries, keys, and values affect the results.

The experimental results are shown in \cref{tab:ablation_results}.
Better performance is achieved on the model which represents an camera image as queries and trajectory planning information as keys and values. The trajectory information is usually simpler and more structured, thus its features are less likely to be disturbed by background noise. Therefore, it is easier for the model to extract parts of them that are highly relevant to the image queries. In this case, the trajectories as values can generate clearer and more focused output features when weighted by the image queries.

In the model where trajectory planning information represents queries and an front camera image does keys and values, even though features of the camera image are highly complex and rich in details, the model cannot fully utilize their richness to produce highly relevant attention distributions due to the simplicity of the trajectory queries. This means that even if the camera image contains a large amount of useful information, the final output features will carry more background noise or irrelevant information due to the limited amount of information available in the trajectory information as queries. Therefore, the model with the trajectory planning information as queries results in a lower performance than the case of the camera image as queries.

\begin{table}[tb]
    \caption{Evaluation results of two ablation methods on two metrics. B4: BLEU-4, RL: ROUGE-L.}
    \label{tab:ablation_results}
    \centering
    \begin{tabular}{cccccccc}
        \toprule
        \multicolumn{2}{c}{Model}
            & \multicolumn{2}{c}{Whole}
            & \multicolumn{2}{c}{Action}
            & \multicolumn{2}{c}{Justification} \\
        \cmidrule(lr){1-2}
        \cmidrule(lr){3-4}
        \cmidrule(lr){5-6}
        \cmidrule(lr){7-8}
           Query & Key, Value & B4 & RL & B4 & R & B4 & RL \\
        \midrule
        Image & Trajectory & \textbf{0.416} & \textbf{0.566} & \textbf{0.483} & \textbf{0.627} & \textbf{0.340} & \textbf{0.535} \\
        Trajectory & Image & 0.367 & 0.539 & 0.409 & 0.582 & 0.307 & 0.514 \\
        \bottomrule
        \end{tabular}
\end{table}

\section{Conclusion}

We discuss interpretability of a driving behavior of the ego vehicle in autonomous driving in order to make its system more trustworthy for its passengers.
We propose a method to generate captions that indicate behaviors of the ego vehicle and their justifications using fused features of trajectory planning information in addition to a front camera image.
Furthermore, we collect a driving dataset which includes planned future trajectories of the ego vehicle, road boundaries, and lane lines with front camera images, and annotated captions at each time frame.
As a result, the proposed models demonstrate better performance on generating precise captions compared to the model without the trajectory planning information.

\par\vfill\par

%
%
\bibliographystyle{splncs04}
\bibliography{main}
\end{document}